# An effective algorithm for hyperparameter optimization of neural networks[1]

G. I. Diaz, A. Fokoue, G. Nannicini, H. Samulowitz

A major challenge in designing neural network (NN) systems is to determine the best structure and parameters for the network given the data for the machine learning problem at hand. Examples of parameters are the number of layers and nodes, the learning rates, and the dropout rates. Typically, these parameters are chosen based on heuristic rules and manually fine-tuned, which may be very time-consuming, because evaluating the performance of a single parametrization of the NN may require several hours. This paper addresses the problem of choosing appropriate parameters for the NN by formulating it as a box-constrained mathematical optimization problem, and applying a derivative-free optimization tool that automatically and effectively searches the parameter space. The optimization tool employs a radial basis function model of the objective function (the prediction accuracy of the NN) to accelerate the discovery of configurations yielding high accuracy. Candidate configurations explored by the algorithm are trained to a small number of epochs, and only the most promising candidates receive full training. The performance of the proposed methodology is assessed on benchmark sets and in the context of predicting drug-drug interactions, showing promising results. The optimization tool used in this paper is open-source.

## Introduction

Data scientists are routinely faced with the task of choosing the best set of parameters for a machine learning model, i.e., a set of parameters that yields the best performance of the predictor on the available dataset. For example, training and testing a neural network (NN) requires determining the structure of the network (number and size of the hidden layers) and of the learning parameters such as learning and dropout rate. Parameters of this type are typically called *hyperparameters*, because they must be determined before the actual training of the model takes place. Only after the training phase will all the parameters of the model, such as the activation thresholds for the neurons, be determined. When the dataset is large, training and testing a single configuration of the hyperparameters may take a long time—several hours or more. Furthermore, choosing values for the hyperparameters that yield high accuracy is sometimes a artful process that typically starts by conforming to heuristic rules, followed by manually fine-tuning the hyperparameters by hand. This results in a long and often tedious process. The aim of this paper is to propose a methodology to automate this task, and to present and discuss a corresponding software implementation. The methodology that we propose is based on formulating the problem of choosing optimal hyperparameters as a box-constrained mathematical optimization problem that has the goal of maximizing the prediction accuracy on a test set, and applying a derivative-free optimization algorithm to such a problem. A derivative-free optimization algorithm is an algorithm that does not require information on the derivatives of any order. A box-constrained problem is an optimization problem defined over a hyperrectangle.

Derivative-free optimization is a well-established area within the field of mathematical optimization, and comprises several techniques—and a comprehensive treatment of the subject is

---

given in [1]. The recent literature contains several works on hyperparameter optimization methodologies applied to automatic configuration of machine learning models, including NNs (e.g., see references [2-6]). Optimization of hyperparameters has also been studied, for example, using a random search (RS) approach [7], Bayesian optimization [8-10], weighted probabilistic extrapolation [11], and other approaches [12,13]. However, to the best of our knowledge, the present paper is the first that applies rigorous derivative-free optimization machinery to this task.

Our paper has the following contributions. First, we discuss how to transform the hyperparameter optimization problem for a NN, which is normally a constrained problem, into a box-constrained problem, and we provide empirical evidence to support our formulation as compared to the most natural formulation. Second, we describe an extension of the derivative-free optimization algorithms implemented in the open-source library RBFOpt [14] to allow parallel, asynchronous evaluations of the function to be optimized, i.e., of the performance of the tested NNs. To the best of our knowledge our approach to parallelize a derivative-free methodology based on radial basis function methods is the first of its kind. The proposed extension is now incorporated in the latest release of RBFOpt, available on GitHub. Third, we provide numerical results that indicate that the hyperparameter optimization approach adopted in this paper can yield significant benefits, and, in particular, it performs better than the popular RS approach.

This paper also has limitations. Most notably, we do not address the problem of choosing the optimal number of training epochs for a NN, and keep it fixed through the hyperparameter optimization process. State-of-the-art Bayesian optimization methods do not have this drawback. We note that RBFOpt allows the exploitation of a computationally cheaper but inaccurate version for the objective function (see [14]), and this capability could be used to accelerate convergence of the hyperparameter optimization algorithm by varying the number of training epochs, but this direction is left for future research. Furthermore, we treat the objective function (accuracy on a validation set) as deterministic, although this not the case. However, in practice we assume that we will never evaluate the same hyperparameter configuration twice, due to limited time availability. Finally, our numerical evaluation is of limited scope, but we note that even with such a limited scope, we can draw some statistically significant conclusions, related to the topics outlined above.

The rest of this paper is organized as follows. In the second section (i.e., the next section), we define the hyperparameter optimization problem. In the third section, we review fundamental concepts in derivative-free optimization. In the fourth section, we discuss how derivative-free optimization can be applied to hyperparameter optimization, more specifically in the context of neural networks, and provide an overview of our approach to parallelize the hyperparameter optimization algorithms implemented in RBFOpt. In the fifth section, we describe our experimental setup and present numerical results. The sixth section concludes the paper.

## The hyperparameter optimization problem

In the context of this paper, a machine learning model is a data-driven mathematical model for the prediction of some property of unseen data. Examples of a property to be predicted are a function value (for regression problems), membership in some class (for classification problems), or relative rankings (for ranking problems). In the rest of this paper, we will use *prediction model* to refer to any of these examples, and we will use the specific case of a NN for a classification problem to illustrate our approach. The structure and training phase of prediction models are governed by a set of hyperparameters. While this set depends on the specific model and corresponding software implementation in use, for a NN typical parameters that are found in most implementations are the number and the size of the hidden layers, the learning rate, and the dropout rate.

Let $X$ describe the set of all valid assignments of values for all the hyperparameters of a prediction model. For example, in the simplest case of all hyperparameters taking unrestricted real values, the set $X$ can be interpreted as a real-valued $n$-dimensional vector space (where $n$ is the number of hyperparameters). Often, though, the hyperparameters of a prediction model take values from different domains; e.g., a learning rate hyperparameter may take only positive real values, a hyperparameter defining the number of hidden layers of a NN, will only be assigned positive integers, and a hyperparameter that defines the type of kernel function for a support vector machine would take categorical values. There may even be implicit constraints among hyperparameter values (e.g., if a given hidden layer in a NN is empty, all subsequent layers must be empty). This implies that the set $X$ usually has a rich structure, and this increases the problem difficulty.

Given a point $\mathbf{x} \in X$, a prediction model can be trained on the available dataset, and its accuracy on unseen data points can be estimated. Typically, this is done by reserving part of the available dataset for validation, or using a $k$-fold cross-validation scheme for some integer $k$. Let $D$ be the dataset. Let $f(\mathbf{x}, D)$ be the estimated performance of a NN parametrized with $\mathbf{x} \in X$ over the dataset $D$, according to the chosen performance metric. In this paper, since we are mostly working with classification problems, we use the accuracy (fraction of correctly classified data points) as the performance metric, but in principle any reasonable metric can be used. The function $f$ will also be called the objective function. The problem of determining the best hyperparameters can then be formulated as follows, where the label *(HPO)* stands for "hyperparameter optimization":

$$\underset{\mathbf{x}}{\mathrm{argmax}}\, \{f(\mathbf{x}, D) : \mathbf{x} \in X\}. \quad (HPO)$$

Notice that the mathematical expression of the function $f$ in closed form may be unknown, but it is a stochastic computable function; in other words, even though we may be unable to provide an analytical expression for $f$, the function is computable given $\mathbf{x}$ and $D$. However, computing $f$ at a single point may require a large amount of time, because it involves training a prediction model on a possibly large dataset, and testing its performance on a validation set.

## Derivative-free optimization: overview

Many applications in engineering and in other fields require the maximization of a performance criterion for which a description in analytical form is not available. This rules out the utilization of classical numerical optimization approaches such as algorithms based on gradient descent [15], because information on the first or higher-order derivatives cannot be computed analytically. In some cases, first and second order derivatives can be numerically estimated by finite differences, but doing so requires a large number of function evaluations: for instance, estimating the gradient of a function defined on $n$-dimensional space at one point requires evaluating the function at $n + 1$ points at least. When the evaluation of the objective function is computationally expensive, as is the case for the application discussed in this paper, such an approach is obviously not viable in practice, and a different methodology becomes necessary. Derivative-free optimization is the area of mathematical optimization dedicated to the task of devising optimization approaches that rely on zero-order information exclusively, i.e. function values. Derivative-free optimization comprises several different methodologies, and we refer the reader to [1] for an overview. Because, in our problem, the evaluation of *f* is particularly computationally expensive and may require several hours of CPU time, we are interested in approaches that try to optimize *f* while performing a small number of function evaluations. This is in contrast to meta-heuristic approaches such as genetic algorithms that rely on zero-order information only but typically perform thousands of function evaluations, resulting in prohibitive computing times. We now briefly describe one of the approaches implemented in the open-source library RBFOpt [14], which as of version 2.0 provides an implementation of two derivative-free optimization algorithms based on surrogate models. The approach that we describe, and that we use in our numerical experiments, is based on the metric stochastic response surface method of [16], with several modifications. The algorithm can be applied to a problem of this form, where the label *(DFO)* stands for "derivative-free optimization":

$$\underset{\mathbf{x}}{\mathrm{argmax}}\ \{f(\mathbf{x}) : \mathbf{x} \in [\mathbf{x}^L, \mathbf{x}^U] \subset \mathbf{R}^n, \mathbf{x}_i \text{ integer } \forall i \in I\}, \quad (DFO)$$

where $\mathbf{x}^L, \mathbf{x}^U$ are vectors of lower and upper bounds for the decision variables, and $I \subseteq \{1..n\}$ is a set of indices of decision variables constrained to take on integer variables only. We note that in [14] and in the software implementation, the problem is assumed to be in minimization form, but here we discuss a maximization problem for consistency and without loss of generality, since a maximization problem can be transformed into a minimization problem by negating the objective function. Notice that *f* is optimized over a domain described by simple lower and upper bounding constraints, i.e. a hyperrectangle, and some decision variables are restricted to integer values. Thus, this is a box-constrained problem with integrality constraints. Other types of constraints are not allowed in this formulation. The vast majority of derivative-free optimization algorithms only apply to box-constrained problems, due to the difficulty of handling general constraints [1].

The algorithm relies on a surrogate model of the objective function *f*, and select new points to evaluate, trying to balance the exploration of unknown parts of the domain of *f*, and the exploitation of the surrogate model, to identify a potential global optimum. Several types of surrogate models are possible, but by default RBFOpt uses a radial basis function model with thin plate splines, combined with a polynomial tail of degree 1 [17]. The algorithm starts by evaluating *f* at $n + 1$ points selected according to a randomly generated Latin Hypercube design within the domain. Let *S* be the set of points at which *f* has been evaluated. Then, at every iteration, the algorithm fits a surrogate model for *f* that interpolates the points in *S*, and chooses the next evaluation point **y** according to two criteria, both of which are to be maximized: the value of the surrogate model at **y** and the Euclidean distance from **y** to the closest point in *S*. A bi-objective optimization problem has in general a possibly infinite set of Pareto-optimal points, but since we are interested in determining a single point **y**, the two objectives are normalized and reduced to a single objective by considering a weighted combination with a weight *w* that determines the tradeoff. More specifically, the weight *w* is chosen according to a cyclic strategy that oscillates between favoring the maximum minimum distance criterion, emphasizing exploration of unknown parts of the domain, and favoring the surrogate model value criterion, emphasizing the choice of points that are supposed to have a large value for $f$ (for a maximization problem) according to the surrogate model. The resulting single-objective optimization problem for the choice of **y** can be solved in several ways: the authors of [16] recommend random sampling, but RBFOpt employs a simple genetic algorithm by default. We note that the genetic algorithm is only applied to the auxiliary optimization problem that determines the choice of **y**, not to the original problem *(HPO)*—while evaluating *f* in *(HPO)* is expensive, evaluating the value of the surrogate model and the minimum distance from the points in *S* is computationally cheap; therefore, the application of the genetic algorithm typically requires only fractions of a second. Once **y** is determined, *f* is evaluated at **y**, the point is added to *S*, and the iteration is complete. The algorithm iterates following this scheme until a stopping criterion is satisfied, typically based on a maximum allowed number of evaluations of *f*, or on a maximum CPU time.

## Derivative-free optimization applied to hyperparameter optimization

We now discuss how we tailored the optimization algorithm of RBFOpt for the solution of *(HPO)* in the specific case of a NN. There is one main difficulty to overcome: when optimizing the hyperparameters of a NN, including the parameters that determine its architecture, the set *X* in *(HPO)* contains other constraints than simple box constraints, because we must enforce that the hyperparameters describe a valid architecture. In particular, there cannot be empty hidden layers of the network between two non-empty hidden layers. This issue can be resolved by adjusting the mathematical formulation of the problem, and this is discussed in the next subsection.

### *Hyperparameter optimization of NNs as a box-constrained problem*

To map *(HPO)* for NNs into the box-constrained *(DFO)* we considered several approaches, and based on our experience in similar contexts, we decided to proceed as follows. Let *u* be an upper bound to the number of hidden layers of the NN that we are willing to consider; *u* could be set to a very large value so as not to restrict the search space, but in practice, a machine learning practitioner can easily proviode a reasonable upper bound for the problem at hand, and *u* will typically be a small number. Similarly, let *l* be the maximum size of a hidden layer that we are willing to consider in terms of number of nodes. In the formulation for *(DFO)*, we utilize *u* + 1 decision variables to determine the size of the hidden layers, say $\mathbf{x}_1..\mathbf{x}_{u+1}$ without loss of generality. The decision variable $\mathbf{x}_{u+1}$ is constrained to be an integer in [0, *u*], and it is used to determine the number of active hidden layers. The remaining variables $\mathbf{x}_1..\mathbf{x}_u$ are constrained to be integers in [1, *l*], and the size of the hidden layers are indicated by taking the first $x_{u+1}$ variables. For example, if *u* = 3 and $(\mathbf{x}_1,..,\mathbf{x}_6) = (20, 10, 30, 10, 40, 50)$, we would construct a NN with 3 hidden layers of size 20, 10, and 30, respectively. Notice that the mapping from the values of $\mathbf{x}_1..\mathbf{x}_{u+1}$ to the structure of the NN introduces symmetry: several values of the decision variables correspond to the same NN structure, and all values of $\mathbf{x}_1..\mathbf{x}_{u+1}$ corresponding to the same structure can be obtained by certain permutations of the decision variables. Symmetry is in general very harmful for optimization algorithms that aim to find the global optimum of a problem (e.g., see [18]). However, a derivative-free approach cannot give global optimality guarantees in finite time, and simply aims to find a solution with large objective function value within few objective function evaluations. For these reasons, symmetry is not necessarily harmful for the problem studied in this paper, and our computational experiments show that the approach that we propose is successful in practice, despite the symmetry.

We briefly explain the rationale of our approach. The variable $\mathbf{x}_{u+1}$ ensures that the design space is explored more uniformly: a naïve formulation that simply bounds the variables $\mathbf{x}_1..\mathbf{x}_u$ between [0, *l*] without imposing an explicit bound on the number of nonempty hidden layers would lead to very dense NNs, because hyperparameter optimization algorithms tend to spread the tested configurations over the entire design space. As a consequence, very few tested configurations would have $\mathbf{x}_i = 0$ for $i = 1,...,u$. Unless there is reason to believe that dense NNs would be more effective on the problem at hand, the formulation proposed above is likely to yield higher accuracy for similar or shorter computing times than the naive formulation. Empirical results, reported in our computational evaluation section, support this claim.

## *Parallelization of RBFOpt*

The evaluation of the objective function *f* is by far the most time-consuming operation in the solution of *(HPO)*: training and assessing the performance of a single NN can take several hours, and for this reason we extended the optimization algorithm implemented in RBFOpt to perform this task in parallel. Notice that the training time of a prediction model such as a NN is not a deterministic quantity, and this introduces difficulty in the parallelization—indeed, many parallel derivative-free optimization algorithms make the simplifying assumption of synchronous parallel objective function evaluations (e.g., see. [19, 20]), but our approach, described next, allows them to be asynchronous. The idea is as follows. We keep a queue of tasks to be performed. These tasks can be of two types—the first type is the evaluation of the objective function *f* at a new point, and the second type is the computation of a new search point at which the objective function must be evaluated. As long as there are available processors, a task is removed from the queue and assigned to a processor. Because of their longer computing time, tasks of type 1 always have priority over tasks of type 2. Within tasks of the same type, we follow a first come, first served policy. Whenever a task of type 1 is completed, it yields a new interpolation point that is added to the set *S*. Whenever a task of type 2 is completed, we check if the newly determined search point is to be discarded because of several criteria also employed in the serial version of the optimization algorithm (see [14]), and if the search point is accepted, we queue a task of type 1 to evaluate *f* at it. An undesirable event can in principle occur if, while *f* is being evaluated at a point **y**, the same point **y** is generated as a search point by concurrent tasks of type 2, and therefore the evaluation of *f*(**y**) is performed multiple times. Clearly, this must be avoided. To this end, whenever a task of type 1 is submitted for processing, we add to *S* a temporary interpolation node at **y** with an objective function value determined by the value of the existing surrogate model at **y**. Since the next search point can never coincide with an interpolation node, even in the serial version of the algorithm (a minimum distance from existing interpolation nodes is required), we are guaranteed that new search points will be distinct. The temporary interpolation node is removed as soon as the associated task of type 1 is completed. It is important to note that several decisions taken by the optimization algorithm depend on the difference between the largest and smallest function value among the interpolation nodes; see [14] for details. The addition of a temporary interpolation node that extends the range of known function values could alter these decisions, and this is especially risky if it is the result of an inaccurate surrogate model with large oscillations. Therefore, we do not allow temporary interpolation nodes to extend the range between known function values: this is achieved by clipping the surrogate model value of a temporary interpolation node to the range of existing function values.

## Computational experiments

We now describe in more detail the empirical evaluation carried out in this study. We first discuss experiments on a standard benchmark dataset (Modified National Institute of Standards and Technology dataset, MNIST) with publicly available code, and then we describe a specific application that uses proprietary code and that is assessed on both publicly available datasets (the WordNet and FreeBase datasets) and non-public datasets.

*Experiments on the MNIST dataset*

To provide a comparison of the performance of RBFOpt with that of popular existing hyperparameter optimization methodologies, we perform a case study on the optimization of a convolutional NN on the well-known and publicly available MNIST hand-written digit recognition dataset, which is a classification problem with 10 classes. The NN is implemented using the Caffe framework and trained using stochastic gradient descent. The search space defined for this instance of (*HPO*) corresponds to NNs with up to 2 convolutional layers with a 5×5 filter size, each one followed by a pooling layer with a 2×2 filter size and stride 2, and up to 2 fully connected layers (called "Inner Product" layers in Caffe). The top level of the NN always consists of 10 units, one for each class, and the winning class is determined applying a softmax. The number of units in each convolutional layer and each fully connected layer varies between 10 and 500 in multiples of 10, whenever the layer is present. We use the model described earlier in this paper whereby there are two additional decision variables to determine the number of hidden layers to be used for each type of layer (convolutional/fully connected). A further hyperparameter determines if the fully connected layers use a rectified linear unit, a linear unit, or a sigmoid activation function. All hyperparameters indicated above are discrete. The remaining hyperparameters are continuous and are the base learning rate (between $10^{-4}$ and $10^{-1}$), the momentum (between 0.05 and 0.95), the weight decay rate (between $10^{-5}$ and $10^{-1}$), and the gamma (between 0.05 and 0.95). The base learning rate and the weight decay rate are on a log scale; in other words, the corresponding decision variables in (*HPO*) represent the $\log_{10}$ of the values used to train the NN, rather than the values themselves: this ensures more effective exploration of the hyperparameter space. The log scale is used to sample these two hyperparameters in RS as well, following [7]. Experiments in this section are performed on IBM's cloud using a server with 16 Intel Xeon CPU E5-2683 v3 clocked at 2.00 GHz, and 32 GB RAM, running Linux.

We compare RBFOpt to RS and Sequential Model-based Algorithm Configuration (SMAC) [21], using the same hyperparameter space for all methodologies. Each algorithm is executed with 10 different random seeds, and the same random seeds are used to initialize the stochastic gradient descent algorithm in the training phase. All algorithms are given the same sequence of random seeds. The MNIST dataset consists of a training set and a test set. We further split the training set into train data (2/3 of the data) and validation data (1/3 of the data). As the objective function in *(HPO)*, we use the best accuracy recorded on the validation set within the first 200 training epochs, with a test interval of 10 epochs. The number of training epochs was chosen after preliminary testing in which we observed that 200 was a good trade-off between speed and accuracy on the validation set. Each run of each algorithm is allowed to explore exactly 100 different hyperparameter configurations, which is equivalent to a budget of 100 objective function evaluations, and no time limit is given. After the budget is depleted, the best configuration found (in terms of accuracy on the validation set) is declared the winner, and we assess its performance on the (blind) test set. For this final test, we first determine the number of training epochs for each configuration using the previous 2/3-1/3 split between train and validation set, and allow up to 1000 epochs to determine the number of epochs that yields the best accuracy on the validation set. Then, we fix the number of epochs and perform the final training and testing. In all our experiments, training is stopped early if at any training epoch after the first 100, the best accuracy on the validation set was achieved before the current number of epochs divided by two; this criterion was also adopted in [7].

Results are summarized in **Table 1** and **Table 2**. Table 1 reports accuracy on the test set for the best configuration discovered, following the procedure described above, while Table 2 reports snapshots of the accuracy on the validation set after 25, 50, 75, 100 iterations. We report the average and standard deviation of the accuracy, and for all pairs of algorithms, how many times each algorithm produces accuracy at least as good as the other, and whether there is a statistically significant difference as detected by a Friedman test followed by post-hoc analysis with a confidence level of 95%. These tables should be interpreted as follows: a number indicates how many times, out of the 10 runs, the algorithm on the row was at least as accurate as the algorithm on the column. Furthermore, if the Friedman test detected a significant difference, we report the direction of such difference; e.g., ">" indicates that the algorithm on the row achieves larger accuracy than the one on the column, with a significance of 95%. The results clearly show that a difference between the algorithms emerges as early as after 50 iterations: both RBFOpt and SMAC perform better than RS, and this trend continues for the rest of the optimization process. Results on the test set are consistent with those on the validation set. RBFOpt achieves slightly higher average accuracy than the other methods, but overall it is only marginally better than SMAC, and no statistically significant difference is detected. Notice that as the number of different runs is only 10, differences are detected only if an algorithm clearly dominates another. We note that during the hyperparameter optimization phase as well as during the final training/test phase, we sometimes encountered numerical troubles: occasionally, the stochastic gradient descent algorithm implemented in Caffe did not converge, and the training error increased until the algorithm aborted. This issue could not be ascribed to any specific reason and occurred with similar frequency with the three tested hyperparameter optimization algorithms. Unfortunately we encountered this issue in the final testing phase of one of the best configurations discovered by RBFOpt, and one discovered by SMAC. We exclude the corresponding runs for all algorithms in the results reported in Table 1, but we verified that this does not change the results of the Friedman test and hence would not change the ranking of the algorithms.

We conclude this section with a brief comment on the performance of the specific unconstrained formulation chosen to map *(HPO)* into *(DFO)*. As previously discussed, we empirically tested the proposed formulation with the naïve formulation that does not employ a decision variable to determine the number of nonempty hidden layers. We used the same framework as in the experiments above, employing RBFOpt to solve *(HPO)*. In terms of accuracy on the validation test after 100 iterations, the naïve formulation and the proposed formulations perform equally: each of them achieves higher accuracy on 5 out of 10 runs. However, the CPU time for the optimization process of the naïve formulation requires a CPU time that is on average 3.6 times larger than the proposed formulation—this is because the optimization algorithm explores mostly dense NN. Since there is no statistically significant advantage in terms of validation score, and the time requirement is larger, our results suggest that the naïve formulation should not be employed.

### *Description of the application to drug-drug interaction*

Aside from the results in the previous section, our hyperparameter optimization algorithm is also applied to a software system called Tiresias [22] that was built to receive various sources of drug-related data and knowledge as inputs, and provide drug-drug interaction (DDI) predictions as outputs. DDIs are a major cause of preventable adverse drug reactions, causing a significant burden on the patients' health and the healthcare system [23]. Tiresias was constructed with a semantic integration of data originating from a variety of sources (including Drugbank, Uniprot, National Drug File – Reference Terminology), followed by building a large knowledge graph containing relations between drugs, diseases, and genes/proteins, and computing several similarity measures between all the drugs. The resulting similarity metrics are used as features in a large-scale logistic regression model to predict potential DDIs. In previous work [22], the similarity features were limited in two ways. First, for the important chemical structure based similarity measures, after computing the fingerprints (i.e. a bit vector capturing the presence of specific substructures) of two drugs, their similarity measures were computed as the cosine similarity of their fingerprints. Second, each similarity measure was constructed by examining only the local neighborhood of each drug in the knowledge graph. We now describe how these limitations were addressed in our present work. This is important because in light of the extensions introduced here, the Tiresias framework can also be applied on some publicly available benchmarks concerning knowledge graphs, and numerical results on these benchmarks will be discussed.

We address the first limitation of [22] by exploring the use of a multi-layer perceptron to directly predict DDIs given the fingerprints of two drugs. The hyperparameters include: number of hidden layers, number of units in each hidden layer, unit activation function (sigmoid, relu, etc), whether pre-training is enabled, and learning parameters (pre-training learning rate, fine tuning learning rate, and Nesterov momentum update parameter). We address the second limitation relying on graph embedding models, which provide dense representations of entities and relations in a knowledge base. The output of such models can be used to define additional similarity metrics, or they may be used directly for link prediction tasks such as that of predicting DDIs. Graph embedding models are trained by minimizing a global cost function in that the entire knowledge graph is considered. In this way, the dense representation of each entity encodes global information about the graph. Although highly expressive graph embedding models exist, recent translation-based models achieve high quality predictions while being scalable to very large datasets. In this context, we have explored extensions of these models which capture not only the knowledge assertions but also ontology information, which is commonly present in large knowledge graphs. Translational graph models generally depend on hyperparameters that include the type of similarity function used for representing the cost between vectors, learning parameters and rates, sampling sizes, the specific update function to be used to minimize the cost function, and margin parameters. The implementation of these models was done in Python with the Theano framework. In our experiments, we try to optimize all hyperparameters, i.e. those of the multi-layer perceptron and those of the graph embedding models, using the accuracy of the predictions over a validation set as the objective function of *(HPO)*.

While the DDI prediction problem is our target application, we run further tests on a publicly available dataset, so as to compare with the state-of-the-art. Because the Tiresias framework is relatively new and it has not been used for benchmarks in its current version, we do not know which region of the hyperparameter space is likely to contain optimal values, or how other hyperparameter optimization methodologies perform. However, the Tiresias framework described above is capable of handling problem types different from the DDI problem; thus, we perform further experiments on the multi-relational data prediction problem described in [24], which can be described as the problem of predicting missing edges in a knowledge graph and for which we can compare to the results reported in the literature. The datasets are derived from WordNet (WN) and Freebase (FB), two large knowledge bases containing multi-relational data. The WordNet dataset is designed to produce an intuitively usable dictionary and thesaurus, while the Freebase dataset is a knowledge base of general facts, and the corresponding machine learning problem aims to learn relationships between words (more details are given in [24]). The two datasets used in our experiments are called WN and FB15k. For an overview of relational machine learning on knowledge graphs, we refer to the recent survey [25].

## *Numerical results using Tiresias*

All our experiments with Tiresias were conducted on a computer cluster composed of NextScale NX360 M4 machines; each machine has 16-24 cores, 256 GB RAM, and 2x Tesla K40 GPUs. Compute jobs are deployed via submission to the Platform Load Sharing Facility. Note that because the experiments are run on a cluster with several concurrent jobs, evaluating the exact computing time is difficult because this depends on the load of the machines. We first evaluate the performance of the hyperparameter optimization methodology on the WN and FB15k datasets of [16] employing the same methodology used in that paper. To test the performance on a single validation point, which consists of a triplet of words, one element of the triplet is removed, and replaced by all possible entities in the dictionary. These newly constructed triplets are then ranked by the prediction mechanism, based on a similarity score. Following [24], we report the proportion of correct entities ranked in the top 10, labeled hits@10, as well as the mean rank of the correct entities. A higher percentage of hits@10 and a lower mean rank indicate better performance. To account for the fact that some of the newly constructed triplets may end up being valid although they were not originally part of the validation set, the authors of [24] propose a filtering method that address the issue. Thus, we report both *raw* results and *filtered* results, as in [24]. As discussed earlier, we only allow the optimizer to perform a small number of training epochs to increase speed. In this benchmark, we allow 10% of the 1000 epochs subsequently used for training.

**Figure 1** shows the progress of the parameter search on the multi-relational data prediction problem of the WN data set. The blue curve indicates performance on the training set, and the red curve indicates performance on the validation (i.e. held-out) set. Within less than 30 iterations, there is a clear jump in performance, and there is an additional one at about 70 iterations. The oscillating behavior is due to the cyclic search strategy discussed earlier in the description of the methodology implemented by RBFOpt, whereby we oscillate between exploration and exploitation. The resulting best configuration obtained by running the training phase for 100 epochs is then trained using 1000 epochs, and evaluated on a test set. We compare the performance of the hyperparameter configuration found by our automatic approach with the performance of a configuration obtained by manually tuning the best configuration reported in [24], and with the best methodology reported in [24], called TransE. We note that in [24], the best configuration is determined in a heuristic way, and we do not know exactly the required effort, however it is reasonable to assume that a manually tuned version of the best configuration in [24] is representative of the performance that can be obtained with a manual or heuristic hyperparameter optimization procedure. Results are reported in **Table 3** using the performance metrics mentioned above. While the obtained performance in terms of hits@10 does not match the results obtained by manually tuning (76.8% vs. 79.9%), the mean rank performance is vastly superior (235 vs 413), indicating that in this particular instance our approach is at least competitive, if not better, than manually tuning the hyperparameters. Furthermore, the performance of the configuration found by RBFOpt + Tiresias is significantly better than the best results reported in [24], which was previously considered the state of the art, both in terms of mean rank and in terms of hits@10. A slightly different scenario occurs on the second data set FB15k—in this case, the best hyperparameter configuration reported in [24] leads to numerical problems in the training phase of Tiresias despite attempts at manually tuning it. On the other hand, RBFOpt discovers a configuration with a performance that is at least comparable to TransE as reported in [24]—it is once again significantly better on the performance metrics for the raw results (mean rank 168 vs. 243, hits@10 47.9% vs. 34.9%), but on the filtered results, [24] reports a better mean rank than RBFOpt + Tiresias, although our approach yields a much larger percentage of hits@10.

**Figure 2** shows the impact of parallelism (using up to 8 parallel evaluations) on the hyperparameter optimization. One can clearly observe how the stages of the search are less pronounced due to the asynchronous runs that break the periodicity of the cyclic search strategy. However, a general upward trend in performance with increasing number of iterations is apparent, and the best configuration determined in the parallel search is comparable to the one determined in the sequential run: 76.6% accuracy and 290 mean rank (because of the randomness involved in the search process, some variability in the results is expected). On the other hand, the parallel run is much faster than the serial run—while the serial run required approximately 92 hours and evaluated ≈130 configurations, the parallel run required approximately 19 hours and evaluated ≈160 configurations. Unfortunately, due to different loads of the computing cluster, we cannot report more precise timing statistics. However, to more accurately estimate the speed improvement that can be achieved by the parallel run, we report statistics on the benchmark set of [14] on a cluster of identical machines with 8 CPUs each and no user-submitted jobs other than ours. The benchmark consists in finding the global optimum of 36 instances of *(DFO)* coming from various sources (see [14] for details), and the location and value of the true optimum of each function are known but are kept hidden from the optimization algorithm. For this experiment, each optimization run was stopped if the number of function evaluations exceeded $60(n + 1)$, where *n* is the dimension of the problem *(DFO)*. Each objective function evaluation took a CPU time drawn uniformly at random between 5 and 10 seconds, and we perform 20 optimization runs with different random seeds on each problem instance, yielding 720 runs in total. In **Table 4** we report, for a given convergence tolerance (i.e. relative distance from the objective function value of the true optimum), the number of problem instances on which the algorithm converged up to the specified tolerance, the average wall-clock time on the instances that were solved by all methods (i.e. with any number of CPUs), and the speedup relative to the serial run. The results indicate a sublinear but still significant speedup, up to more than a factor 3 when 8 CPUs are used in parallel, although unsurprisingly the "price to pay" is that the number of solved instances is smaller.

Finally, we report results on the DDI prediction problem. For this instance of *(HPO)*, we aim to construct a NN with up to 4 hidden layers with up to 5000 units each. Unfortunately there is a lack of a well-established baseline for this method, because the extension of the Tiresias framework discussed in this paper has not been previously tested. In **Figure 3** we report the accuracy on the validation set. Figure 3 shows the progress of the hyperparameter optimization algorithm when training for 35 epochs. While there exist some initial improvements, obtaining validation scores larger than 59% appears to be difficult, and it is only achieved more frequently at the later stages of the search. Next, we use the best configuration discovered on the validation set and train it up to 100 epochs, comparing the resulting accuracy on a test set with the accuracy of a hyperparameter configuration chosen by domain experts in a heuristic way. One main difference in the two configurations is that the user-generated parameterization only uses one hidden layer while two are selected by RBFOpt. The two resulting accuracies are 53.59% (optimized configuration) and 47.35% (domain expert), showing a substantial improvement. While this benchmark is only marginally significant, because it does not include a comparison to other rigorous hyperparameter optimization methodologies, it serves the purpose of showing that in a practical application, our approach quickly and successfully finds a configuration with higher accuracy than what would have been guessed by domain experts, while being fully automated.

**Conclusion**

We presented a methodology for the hyperparameter optimization problem and a corresponding software implementation. The hyperparameter optimization problem consists of finding values of the parameters of a machine learning algorithm to achieve the largest prediction accuracy on a dataset. Our methodology is based on casting the problem as a derivative-free optimization problem, and solving it as such. While the methodology cannot be guaranteed to find the global optimum in a finite number of iterations, we show empirically that it finds values of the hyperparameters yielding higher prediction accuracy than those determined by a domain expert. Moreover, on a set of benchmark instances from the literature, we obtain results that are at least comparable, and sometimes better, than popular existing algorithms such as random search and sequential model-based algorithm configuration. This suggests that using derivative-free optimization techniques for the hyperparameter optimization problem in machine learning is a promising approach that has the potential to have an impact in practice. In future research, we will try to address the problem of choosing the number of training epochs, combining the approach discussed in this paper with learning-curve-based optimization approaches such as in [11, 26].

**Gonzalo I. Diaz** *Keble College, University of Oxford, Oxford, OX1 3PG UK (gonzalo.diaz@cs.ox.ac.uk).* Mr. Diaz received an M. Sc. degree in Computer Science from the P. Catholic University of Chile in 2013, and is currently a Ph.D. student in the Department of Computer Science at the University of Oxford, where he is working under the supervision of Prof. Michael Benedikt. His research is focused on semantic web technologies and data management. During the summer of 2016, he was an intern at the IBM Thomas J. Watson Research Center.

**Achille Fokoue-Nkoutche** *IBM Research Division, Thomas J. Watson Research Center, Yorktown Heights, NY 10598 USA (achille@us.ibm.com).* Mr. Fokoue-Nkoutche is a Research Staff Member in the Cognitive Computing department at the IBM Thomas J. Watson Research Center. He received his M.S. degree in computer science in 1999 from Ecole Centrale Paris, France. In 2001, he joined IBM at the T. J. Watson Research Center, where his research has focused on knowledge representation and reasoning, the semantic web, and machine learning. In particular, Mr. Fokoue-Nkoutche has developed theories, algorithms and systems for scaling reasoning over large and expressive description logics knowledge-bases that tolerate inconsistencies and uncertainties. He has also applied semantic web technologies and machine learning in a variety of domains ranging from healthcare and life sciences to text analysis. He is author or coauthor of 16 patents, 40 research papers, and the Web Ontology Language version 2.0 (OWL 2.0) W3C standard.

**Giacomo Nannicini** *IBM Research Division, Thomas J. Watson Research Center, Yorktown Heights, NY 10598 USA (nannicini@us.ibm.com).* Dr. Nannicini is a Research Staff Member in the Mathematical Sciences Department at the IBM T. J. Watson Research Center. He received a Ph.D. degree in computer science from Ecole Polytechnique (France) in 2009. Before joining IBM in 2016, Dr. Nannicini had postdoctoral research positions at Carnegie Mellon University and MIT, and he was an assistant professor at the Singapore University of Technology and Design. His main research interest is mathematical optimization broadly defined and its applications. Dr. Nannicini is the recipient of the 2012 Glover-Klingman prize and the 2015 Robert Faure prize.

**Horst Samulowitz** *IBM Research Division, Thomas J. Watson Research Center, Yorktown Heights, NY 10598 USA (samulowitz@us.ibm.com).* Dr. Samulowitz is a Research Staff Member in the Cognitive Computing department at the IBM T. J. Watson Research Center. He received a M.Sc. degree from RWTH Aachen in 2003, and a Ph.D. degree in computer science from the University of Toronto in 2008. He was a postdoctoral researcher at Microsoft Research England for 2 years and a researcher at NICTA and the University of Melbourne. He subsequently joined the IBM T. J. Watson Center, where he is working on projects in the area of artificial intelligence and applied machine learning, including meta-learning, interactive machine learning systems, combinatorial reasoning and optimization, and efficient inference technology for propositional logic and probabilistic reasoning.

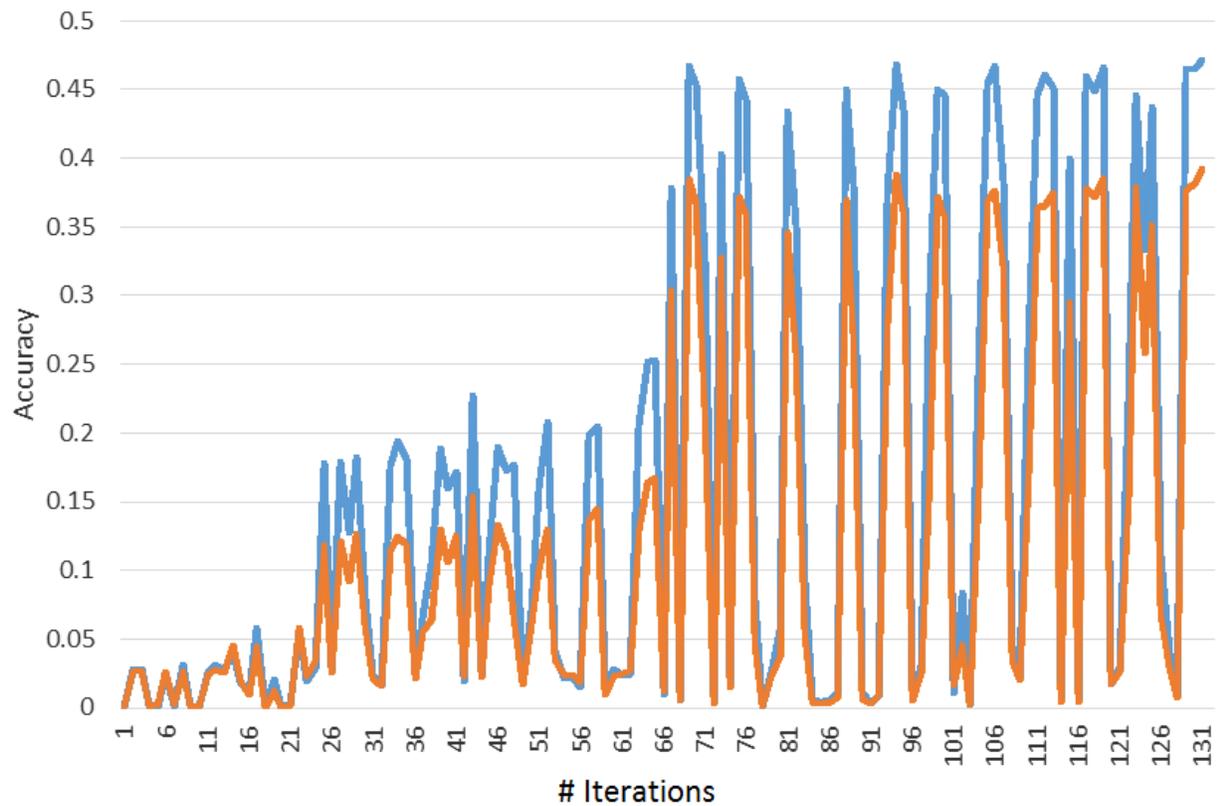

**Figure 1**: Accuracy on the training set (blue curve) and validation set (red curve) for a serial run of RBFOpt + Tiresias.

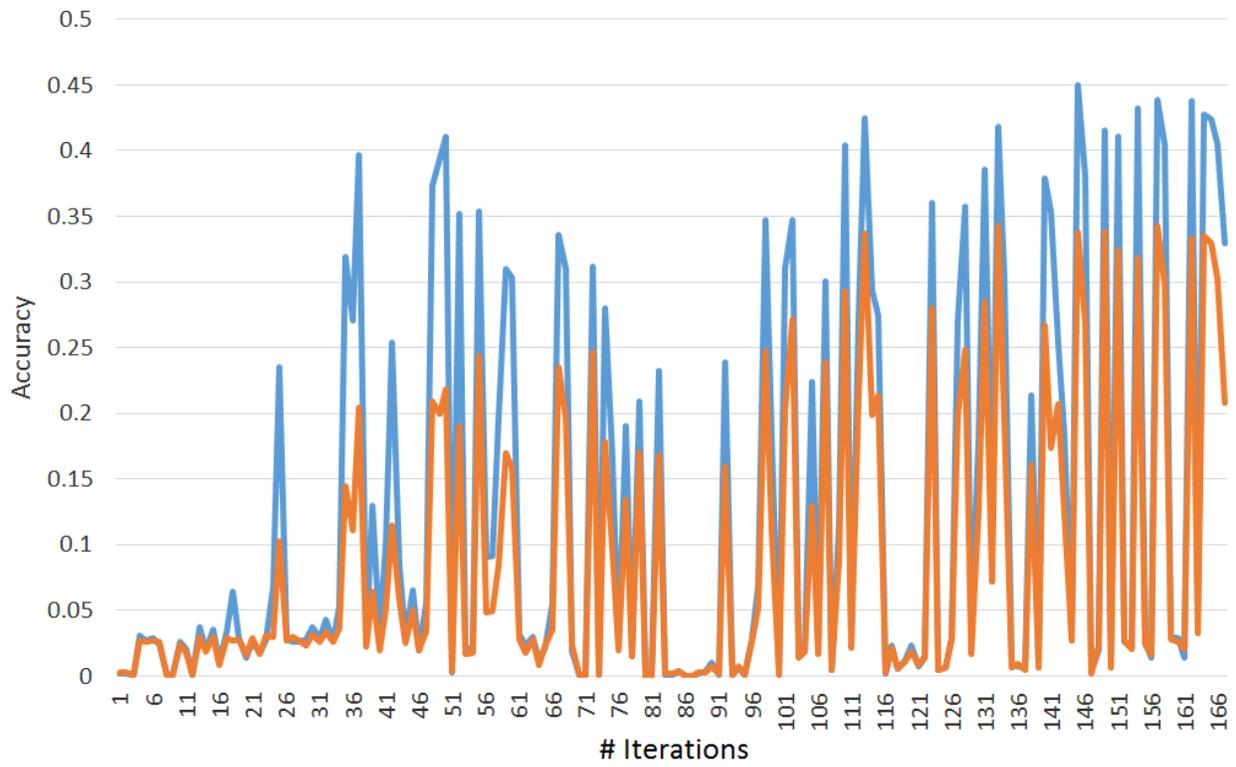

**Figure 2**: Accuracy on the training set (blue curve) and validation set (red curve) for a parallel run of RBFOpt + Tiresias.

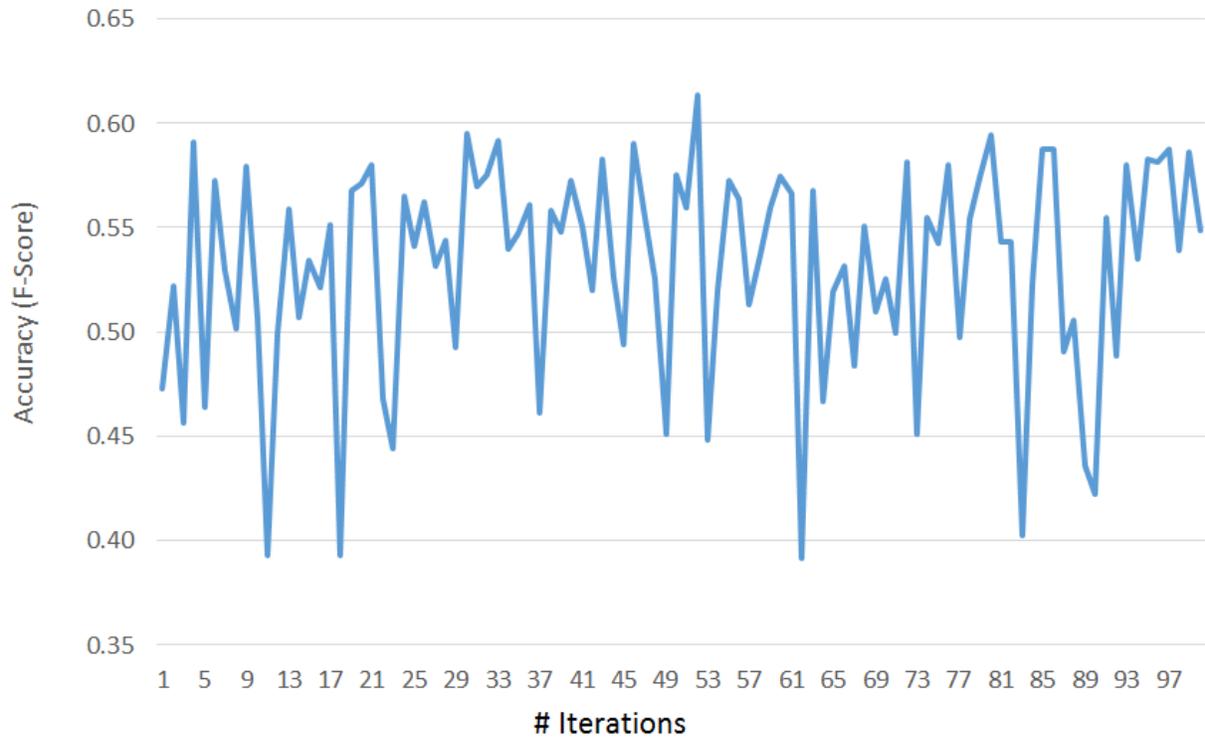

**Figure 3**: Accuracy (F-Score) on the validation set for a sequential run of RBFOpt + DDI.

**Table 1** Results on the MNIST dataset: accuracy of the best configuration on the test set. (Avg. acc: average accuracy; stdev: standard deviation.)

|  | Count better (Friedman test) | | | Avg. acc. (stdev) |
|---|---|---|---|---|
| Algorithm | RBFOpt | RS | SMAC | |
| RBFOpt |  | 8 (>) | 5 (=) | 97.93 (0.01) |
| RS | 0 (<) |  | 1 (<) | 89.07 (0.06) |
| SMAC | 3 (=) | 7 (>) |  | 97.54 (0.01) |

**Table 2** Results on the MNIST dataset: accuracy on the validation set. (RB: RBFOpt; SM: SMAC.)

|  | Iteration 25 | | | | Iteration 50 | | | | Iteration 75 | | | | Iteration 100 | | | |
|---|---|---|---|---|---|---|---|---|---|---|---|---|---|---|---|---|
|  | Count better (Friedman test) | | | Avg acc. (stdev) | Count better (Friedman test) | | | Avg acc. (stdev) | Count better (Friedman test) | | | Avg acc. (stdev) | Count better (Friedman test) | | | Avg acc. (stdev) |
| Algorithm | RB. | RS | SM. |  | RB. | RS | SM. |  | RB. | RS | SM. |  | RB. | RS | SM |  |
| RBFOpt |  | 8 (=) | 5 (=) | 94.70 (1.30) |  | 8 (>) | 5 (=) | 95.01 (1.28) |  | 9 (>) | 4 (=) | 95.44 (0.89) |  | 9 (>) | 6 (=) | 95.66 (0.82) |
| RS | 2 (=) |  | 3 (=) | 93.39 (0.69) | 2 (<) |  | 1 (<) | 93.48 (0.70) | 1 (<) |  | 1 (<) | 93.82 (0.45) | 1 (<) |  | 1 (<) | 94.02 (0.46) |
| SMAC | 5 (=) | 7 (=) |  | 94.19 (1.31) | 5 (=) | 9 (>) |  | 94.86 (0.76) | 6 (=) | 9 (>) |  | 95.43 (0.77) | 4 (=) | 9 (>) |  | 95.48 (0.78) |

**Table 3** Results for the multi-relational data prediction problem.

| Dataset | WN | | | | FB15k | | | |
|---|---|---|---|---|---|---|---|---|
| Metric | Mean Rank | | Hits@10 (%) | | Mean Rank | | Hits@10 (%) | |
| Eval. setting | Raw | Filtered | Raw | Filtered | Raw | Filtered | Raw | Filtered |
| RBFOpt + Tiresias | 235 | 223 | 76.8 | 88.5 | 168 | 146 | 47.9 | 55.3 |
| Manually tuned | 413 | 400 | 79.9 | 93.0 | -- | -- | -- | -- |
| TransE [23] | 263 | 251 | 75.4 | 89.2 | 243 | 125 | 34.9 | 47.1 |

**Table 4** Performance of the parallel version of RBFOpt. Speedup indicates a speedup factor.

| | Convergence to 1% of optimality | | | Convergence to 0.1% of optimality | | |
|---|---|---|---|---|---|---|
| # CPUs | # solved | time (sec) | speedup | # solved | time (sec) | speedup |
| 1 | 551 | 206.5 | 1.00 | 509 | 195.3 | 1.00 |
| 2 | 531 | 130.0 | 1.58 | 501 | 117.1 | 1.66 |
| 4 | 532 | 83.9 | 2.45 | 494 | 81.4 | 2.39 |
| 8 | 499 | 65.3 | 3.16 | 442 | 63.5 | 3.07 |